\documentclass[runningheads]{llncs}
\usepackage[T1]{fontenc}
\usepackage{graphicx,verbatim}
\usepackage{amsmath}
\usepackage{amssymb} 
\usepackage{booktabs}
\usepackage{multirow}
\usepackage{hyperref}
\usepackage{color}
\usepackage{xcolor}

\begin{document}
\title{PanORama: Multiview Consistent Panoptic Segmentation in Operating Rooms}
%
\author{Tuna Gürbüz\thanks{Equal contribution.}\inst{1}, Ege Özsoy\inst{\star,1,2}, Tony Danjun Wang\inst{1,2}, Nassir Navab\inst{1,2}}

\institute{
Computer Aided Medical Procedures, Technische Universit{\"a}t M{\"u}nchen, Germany\\
MCML, Germany
}
  
\maketitle              
\begin{abstract}
Operating rooms (ORs) are cluttered, dynamic, highly occluded environments, where reliable spatial understanding is essential for situational awareness during complex surgical workflows. Achieving spatial understanding for panoptic segmentation from sparse multiview images poses a fundamental challenge, as limited visibility in a subset of views often leads to mispredictions across cameras. To this end, we introduce PanORama, the first panoptic segmentation for the operating room that is multiview-consistent by design. By modeling cross-view interactions at the feature level inside the backbone in a single forward pass, view consistency emerges directly rather than through post-hoc refinement. We evaluate on the MM-OR and 4D-OR datasets, achieving $>70$ \% Panoptic Quality (PQ) performance, and outperforming the previous state of the art. Importantly, PanORama is calibration-free, requiring no camera parameters, and generalizes to unseen camera viewpoints within any multiview configuration at inference time. By substantially enhancing multiview segmentation and, consequently, spatial understanding in the OR, we believe our approach opens new opportunities for surgical perception and assistance. Code will be released upon acceptance.


\end{abstract}

\begin{figure}
    \centering
    \includegraphics[width=\linewidth]{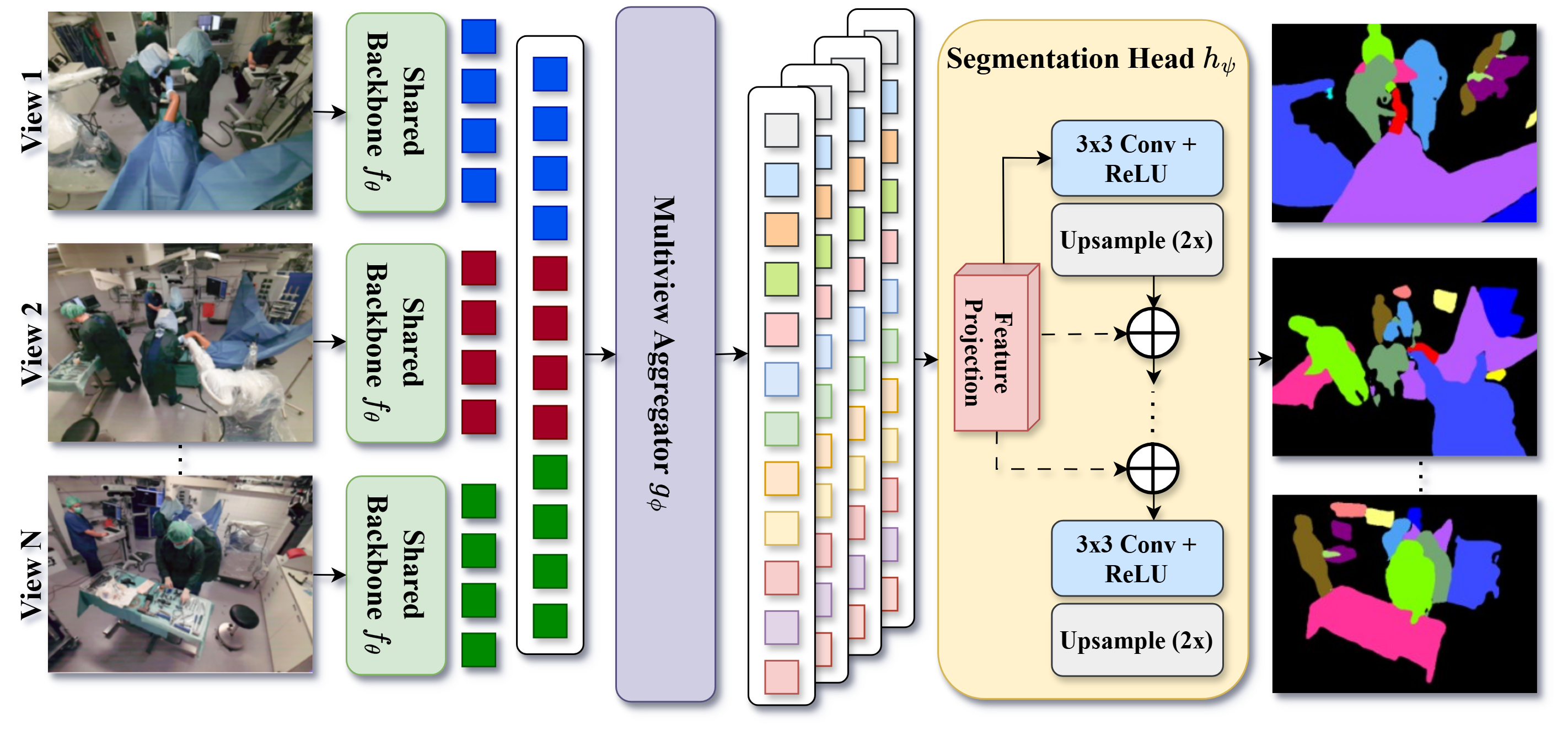}
    \caption{Overview of PanORama. Each input view is independently encoded by a shared visual backbone $f_\theta$, producing per-view patch token embeddings. The multiview aggregator $g_\phi$ fuses these embeddings via local and global attention, enabling cross-view reasoning before mask prediction. The segmentation head $h_\psi$ progressively upsamples the aggregated features through hierarchical convolutional stages with skip connections to produce panoptic masks for each input view in a single forward pass.}
    \label{fig:ModelArchitecture}
\end{figure}

\section{Introduction}
Operating rooms are densely populated and highly occluded workspaces, making reliable spatial understanding a central bottleneck for automated perception and surgical assistance in surgical data science~\cite{lalys2014,maier2017surgical,ozsoy2023holistic,ozsoy2024oracle}. Achieving this spatial intelligence requires understanding what is where, what is occluded, and how entities relate in 3D. This is particularly challenging in the OR, as it can exhibit severe clutter, contain small, transparent or specular materials, and involve frequent interactions. Moreover, viewpoint-dependent visibility means that critical evidence may be absent in one camera but present in another, making spatial understanding inherently a multiview problem. A comprehensive spatial scene understanding must therefore be inherently global, built directly from all partial observations. 

The community has made significant progress towards surgical intelligence, steadily expanding the scope from narrowly defined recognition tasks\cite{twinanda2016endonet,jin2020multi,rendezvous,ding2022carts,sharghi2020automatic} toward richer semantic scene understanding, modeled as question answering \cite{he2024pitvqa,yuan2024advancing,ozsoy2025specialized}, action detection~\cite{fujii2024egosurgery}, and scene graph generation~\cite{ozsoy20224d,ozsoy2023,pei2024s,guo2024tri,ozsoy2024oracle,ozsoy2025mmor,ozsoy2025egoexor}. In contrast, spatial understanding in the OR remains underexplored. To date, 
SegmentOR~\cite{bastian2023segmentor} is one of the very few works that tackles segmentation in OR, by operating on fused 4D point clouds and enabling weakly supervised segmentation with substantially reduced labeling effort. However, point-cloud-based representations are inherently sparse and often limited to coarse-grained objects (e.g., operating table, tool table) and are thus unsuitable for fine-grained understanding of small, but crucial objects (e.g., hammer, saw). Complementarily, MM-OR~\cite{ozsoy2025mmor} provides the first publicly available OR panoptic segmentation annotations across multiple camera views and establishes a benchmark using single-view panoptic segmentation as a baseline. Yet, it still leaves open the question of how to obtain a globally coherent segmentation-based scene representation from sparse multiview observations.

Beyond the OR domain, 3D perception progressed rapidly, with recent works successfully showing multiview understanding capability for tasks such as 3D object detection~\cite{wang2022detr3d,liu2024ray,wang2023exploring}, depth estimation~\cite{zhang2024multi}, multiview clustering~\cite{yan2023gcfagg}, and reconstruction~\cite{wang2025vggt,keetha2025mapanything}. However, these advances have not been yet investigated in the OR domain, where sparse overlap, heavy occlusions, and limited data make it difficult to learn globally coherent multiview semantics.

To address this gap, we introduce \textbf{PanORama}, the first OR panoptic segmentation model that is multiview-consistent \emph{by design}. Our key idea is to enable cross-view interaction directly at the feature level before mask decoding, so that view consistency emerges from a single end-to-end forward pass rather than post-hoc processing. Our architecture yields a globally coherent representation, while being calibration-free and supporting an arbitrary number of input views at inference time. In extensive evaluations on MM-OR and 4D-OR, PanORama achieves state-of-the-art panoptic segmentation performance, reaching \textbf{71.85\%} PQ and improving over the DVIS++~\cite{ozsoy2025mmor} baseline by over \textbf{7\%} PQ on MM-OR. Beyond the main benchmark results, we conduct two targeted analyses that probe the core capabilities enabled by multiview feature aggregation, first we show that PanORama can effectively leverage additional \emph{unlabeled} synchronized views during training to further improve performance at test time, and second, we demonstrate substantially improved generalization to \emph{novel} camera viewpoints that were withheld from supervision during training. By learning view-consistent dense semantics directly from sparse multiview images, PanORama paves the way for practical, scalable spatial intelligence in the OR.
\section{Method}

\subsection{Multiview Consistent Panoptic Segmentation}
Our goal is to obtain view-consistent panoptic segmentations that support accurate, fine-grained spatial understanding of the operating room. Given $V$ synchronized RGB images $(I_1,\dots,I_V)$, our goal is to predict a panoptic segmentation for each view while leveraging complementary evidence across views. We model the output for view $v$ via dense prediction, as a semantic score map $S_v \in \mathbb{R}^{H\times W\times C}$ over $C$ classes. We consider the sparse multiview setting, prevalent in the OR due to the large number of clinicians, instruments, and equipment, where views may only partially overlap and objects can be visible in only a subset of cameras. In this setting, multiview-consistent panoptic segmentation aims to produce per-view predictions that are globally coherent across the camera set.

\subsection{Architecture Overview} 
Our model comprises three components: a shared visual backbone, a multiview feature aggregator, and a dense panoptic segmentation head (Fig.~\ref{fig:ModelArchitecture}). Our central insight is that multiview-consistent segmentation benefits from an aggregator that has already learned cross-view fusion with 3D consistency. We introduce a novel reformulation of the VGGT-style~\cite{wang2025vggt} aggregation module, transforming it from a 3D reconstruction mechanism into a multiview consistent cross-view segmentation formulation that explicitly exchanges complementary evidence across cameras. The segmentation head then maps the resulting multiview-aware tokens to per-view dense semantic score maps. This design induces view consistency \emph{by construction} in a single end-to-end forward pass, as cross-view interaction occurs prior to decoding rather than via post-hoc reconciliation.

\paragraph{Shared Visual Backbone.}
Given $V$ synchronized RGB images $(I_1,\dots,I_V)$, the shared backbone $f_{\theta}$ processes each view independently using shared weights and extracts patch-level token embeddings
\begin{equation}
X_v = f_{\theta}(I_v), \quad X_v \in \mathbb{R}^{N \times D},
\end{equation}
where $N$ is the number of patch tokens and $D$ is the token dimension. The per-view tokens are introduced to preserve local appearance and spatial layout while serving as the
input to the multiview feature aggregator.

\paragraph{Multiview Feature Aggregator.}
The aggregator $g_{\phi}$ fuses complementary evidence across cameras by performing cross-view feature interaction over the ordered tuple of per-view tokens $(X_1,\dots,X_V)$:
\begin{equation}
(\tilde{X}_1,\dots,\tilde{X}_V) = g_{\phi}(X_1,\dots,X_V), \quad \tilde{X}_v \in \mathbb{R}^{N \times D}.
\end{equation}
Concretely, the aggregator concatenates the token sequences from all views and alternates \emph{local} and \emph{global} attention blocks. Here, \emph{local} attention denotes standard self-attention applied independently within each view, i.e., tokens from view $v$ attend only to tokens from the same view (a block-diagonal attention mask). In contrast, \emph{global} attention denotes standard self-attention over the concatenated multiview token sequence, allowing tokens to attend across views and exchange complementary evidence. This enables the model to capture shared scene structure and correspondences under partial overlap and occlusion, without predicting or supervising explicit geometry (e.g., camera pose or depth) and without requiring camera calibration. The output is an ordered tuple of multiview-aware per-view token representations $(\tilde{X}_1,\dots,\tilde{X}_V)$ and intermediate multi-scale features, that are subsequently processed by the panoptic segmentation head to produce per-view panoptic outputs.

\paragraph{Panoptic Segmentation Head.}
Finally, a dense panoptic segmentation head $h_{\psi}$ is introduced to map each multiview-aware representation $\tilde{X}_v$ to per-pixel semantic scores. The head uses a coarse-to-fine decoder. It reshapes the token sequence into a low-resolution feature grid and progressively upsamples it to full resolution. At each stage, convolutional refinement is applied and higher-resolution intermediate features are fused via skip connections to recover spatial detail. The final stage maps them to per-pixel semantic scores.
\begin{equation}
S_v = h_{\psi}(\tilde{X}_v), \quad S_v \in \mathbb{R}^{H \times W \times C}
\end{equation}

\subsection{Training and Inference}
We train PanORama end-to-end with supervision on the annotated views by applying a combination of binary cross-entropy and Dice loss to the per-view score maps $S_v$.

\begin{equation}
\mathcal{L} = \mathcal{L}_{\text{BCE}}(S, Y) + \mathcal{L}_{\text{Dice}}(S, Y),
\end{equation}

To ensure robustness to different multiview configurations, we randomize the view order during training. Notably, at inference time, PanORama accepts an arbitrary number of synchronized RGB views and outputs a panoptic prediction for each input view.

\section{Experiments}

\begin{table*}[t]
\centering
\caption{\textbf{Main results.} Panoptic Quality (PQ$\uparrow$) as percentage on MM-OR and 4D-OR.
Train: MM-OR, Both = MM-OR+4D-OR.
Columns 1/4/8 report PQ over temporal windows of 1, 4, and 8 consecutive frames.}
\label{tab:train_eval_pq_sv_mv}

\setlength{\tabcolsep}{3.5pt}
\renewcommand{\arraystretch}{1.15}
\footnotesize
\begin{tabular}{c c l ccc ccc}
\toprule
\multirow{2}{*}{\textbf{View}} &
\multirow{2}{*}{\textbf{Train}} &
\multirow{2}{*}{\textbf{Method}} &
\multicolumn{3}{c}{\textbf{MM-OR PQ}} &
\multicolumn{3}{c}{\textbf{4D-OR PQ}} \\
\cmidrule(lr){4-6} \cmidrule(lr){7-9}
& & & \textbf{1} & \textbf{4} & \textbf{8} & \textbf{1} & \textbf{4} & \textbf{8} \\
\midrule

\multirow{4}{*}{Single}
& MM-OR & DVIS++~\cite{ozsoy2025mmor} & 64.60 & 59.60 & 59.30 & \multicolumn{3}{c}{--} \\
& Both  & DVIS++~\cite{ozsoy2025mmor} & 64.10 & 62.30 & 61.70 & 64.90 & \textbf{67.80} & \textbf{67.50} \\
& MM-OR & PanORama (Single)           & 68.91 & 66.28 & 64.75 & \multicolumn{3}{c}{--} \\
& Both  & PanORama (Single)           & 69.21  & 66.40  & 64.97  & 62.35  & 59.19  & 56.92  \\
\midrule

\multirow{2}{*}{Multi}
& MM-OR & PanORama (Multi)            & \textbf{71.85} & \textbf{68.80} & \textbf{66.91} & \multicolumn{3}{c}{--} \\
& Both  & PanORama (Multi)            & 71.42 & 68.67 & 66.66 & \textbf{68.08} & 63.52 & 59.14 \\
\bottomrule
\end{tabular}
\end{table*}

\paragraph{Datasets.}
We conduct all experiments on MM-OR~\cite{ozsoy2025mmor} and 4D-OR~\cite{ozsoy20224d} using the official train/val/test splits. MM-OR contains 92{,}983 timepoints of simulated robotic knee replacement procedures recorded from five fixed external RGB viewpoints, where three have segmentation annotations. While the dataset also provides additional modalities, in this work, we only use the synchronized external RGB views for multiview panoptic segmentation. 4D-OR consists of 6{,}734 timepoints of simulated knee replacement surgeries recorded from six fixed external RGB viewpoints, with again three having segmentation annotations. Similar to MM-OR, we use only the synchronized external RGB views and the available panoptic annotations for training and evaluation. Unless stated otherwise, we use all available views as input and apply supervision and evaluation only on the annotated views.

\paragraph{Implementation Details.}
All models are implemented in PyTorch. The backbone and multiview aggregator are initialized from pretrained DINO~\cite{oquab2023dinov2} and VGGT~\cite{wang2025vggt} weights, respectively. We add dropout in the panoptic segmentation decoder to mitigate overfitting. Unless stated otherwise, models are trained on individual timepoints. We optimize with AdamW~\cite{loshchilov2017decoupled} (learning rate $1\times 10^{-5}$, weight decay $0.05$) for 21 epochs with mini-batches of 5--6 samples (dataset-dependent) on a single NVIDIA RTX 3090 (24 GB).

\paragraph{Evaluation Metrics.}
We follow the official MM-OR panoptic segmentation evaluation protocol and report Panoptic Quality (PQ). PQ jointly measures segmentation quality and recognition quality by matching predicted and ground-truth segmentations and combining their overlap (IoU) with detection accuracy, yielding a single score in $[0,100]$ (higher is better). For temporality-aware evaluation, we additionally report PQ over temporal windows of 1 / 4 / 8 consecutive frames, where longer windows evaluate consistency under temporal aggregation.

\begin{figure}[t]
    \centering
    \includegraphics[width=1.0\linewidth]{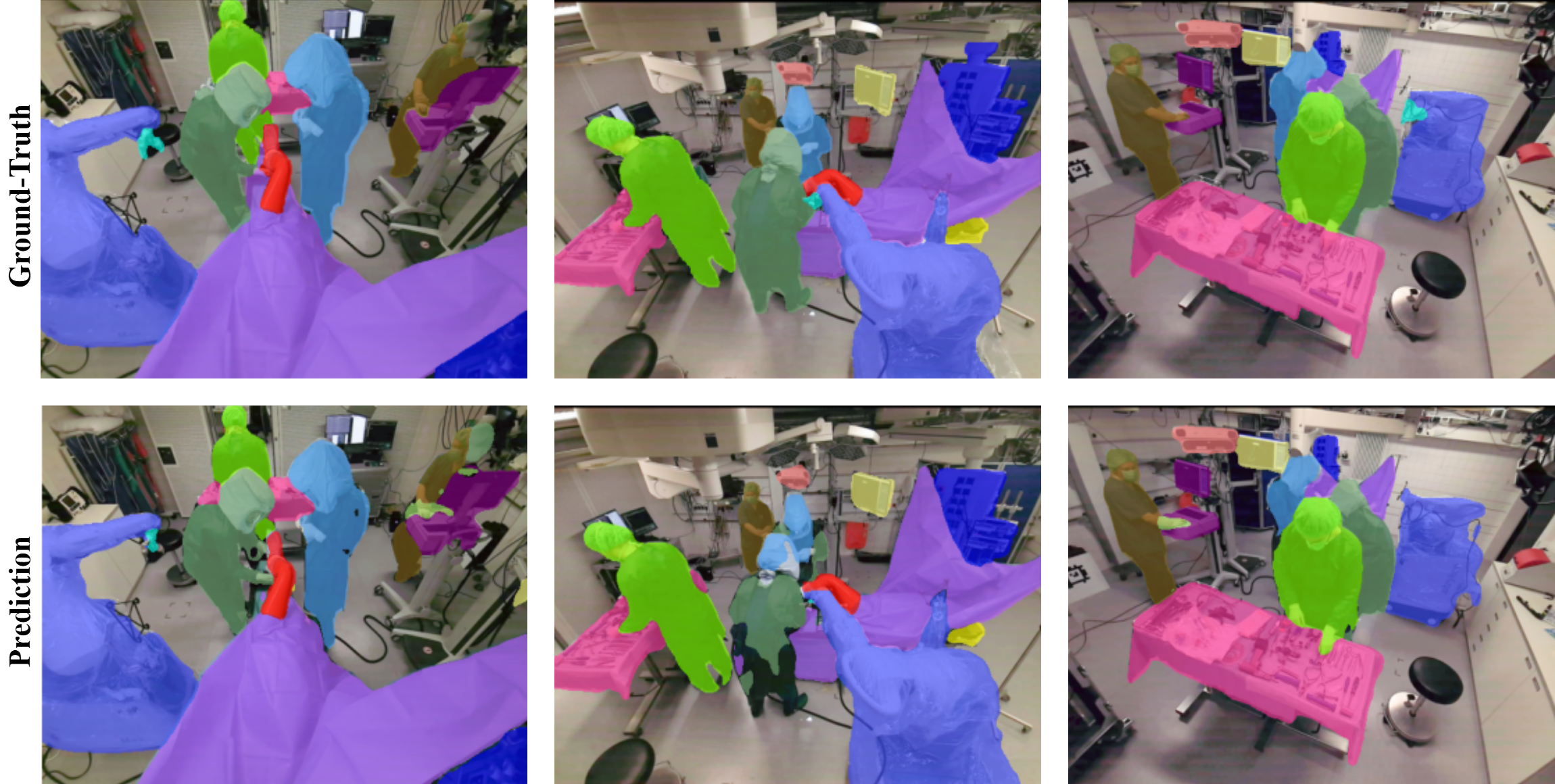}
    \caption{Qualitative panoptic segmentation results on the MM-OR test set. Each column corresponds to one of the five synchronized camera views, with ground-truth annotations (top row) and predicted masks (bottom row) overlaid on the color image.}
    \label{fig:qualitative}
\end{figure}

\begin{figure}[t]
    \centering
    \includegraphics[width=\linewidth]{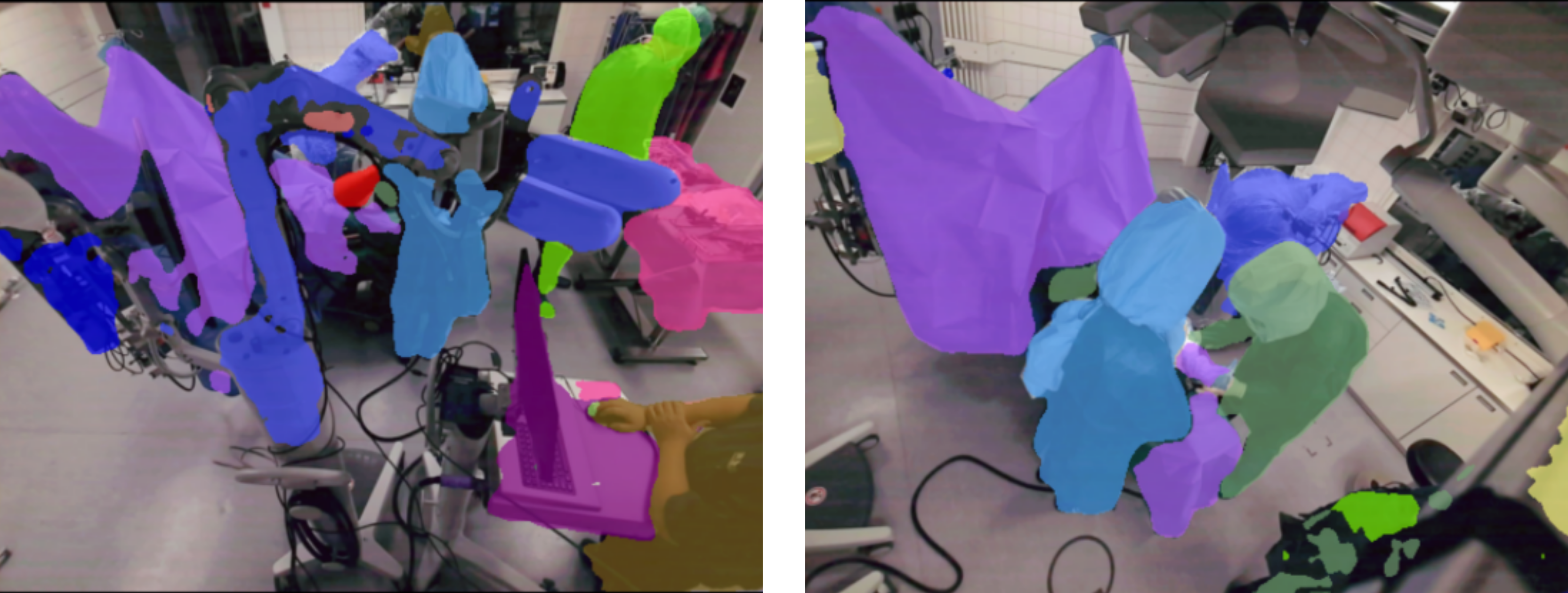}
    \caption{Qualitative panoptic segmentation results on the MM-OR test set for views lacking ground-truth annotations and excluded from supervision. These predictions are produced jointly with the supervised views shown in Fig.~\ref{fig:qualitative} in a single inference pass, yet without any view-specific training signal, demonstrating PanORama's generalization capability within a multiview context.}
    \label{fig:qualitative_unsupervised}
\end{figure}

\section{Results and Discussion}
\paragraph{Main Results.}
Table~\ref{tab:train_eval_pq_sv_mv} summarizes PQ on MM-OR and 4D-OR under different training settings and evaluation protocols. PanORama consistently outperforms the DVIS++ baseline across datasets and temporal window sizes. On MM-OR, PanORama improves by more than 7 \% PQ points for single-frame evaluation (71.85 vs. 64.1) and maintains a clear margin under temporal window aggregation (1/4/8 frames), establishing a new state of the art for OR panoptic segmentation. Figure~\ref{fig:qualitative} shows that PanORama produces globally coherent segmentations across views under occlusions and partial visibility. In particular, when objects are only partially visible in a subset of cameras, PanORama can still yield consistent per-view predictions by aggregating complementary evidence across views. Additional qualitative visualizations are provided in the supplementary video.

\paragraph{Single-view vs. Multiview.}
To isolate the contribution of multiview aggregation, Table~\ref{tab:train_eval_pq_sv_mv} compares a single-view PanORama variant against the full multiview model. While the single-view variant already improves over DVIS++, adding multiview aggregation yields additional gains, confirming that cross-view feature interaction provides complementary spatial evidence beyond stronger per-image modeling.

\paragraph{Leveraging Unlabeled Views.}
A key advantage of PanORama is that it can ingest additional synchronized views even when panoptic annotations are available only for a subset of cameras. Table~\ref{tab:eval_input_views} shows that including two additional unlabeled views during training does not degrade performance on the annotated views and yields further improvements when all views are provided at inference time. This indicates that unlabeled views contribute useful complementary evidence and strengthen the learned multiview representation under sparse supervision.

\paragraph{Novel Camera Views.}
We evaluate generalization to previously unseen camera viewpoints by withholding one camera during training and reintroducing it only at test time. Concretely, we exclude camera~4 from the training inputs; at inference time, camera~4 is provided again and PQ is computed on this previously unseen view. As shown in Table~\ref{tab:missing_view} and Figure~\ref{fig:qualitative_unsupervised}, PanORama generalizes substantially better to the novel view than the single-view DVIS++ baseline, indicating that PanORama learns a globally coherent multiview representation rather than treating each camera completely independently.

\paragraph{Discussion.}
While our current formulation and method focus on per timepoint predictions, without explicitly enforcing temporal consistency, the consistent gains across temporal evaluation windows indicate that multiview aggregation provides robust spatial cues that complement temporality. Nonetheless, we believe an explicit integration of temporal modeling can be a promising future direction, potentially further improving consistency and robustness in long, dynamic surgical workflows.

\begin{table}[t]
\centering
\caption{\textbf{Leveraging unlabeled views.} Effect of the number of input views during training and inference on MM-OR.
Panoptic annotations are available for 3 views only; PQ ($\uparrow$) is computed on the annotated views.}
\label{tab:eval_input_views}

\setlength{\tabcolsep}{6pt}
\renewcommand{\arraystretch}{1.15}
\small
\begin{tabular}{c c c}
\toprule
\textbf{\#Train Views} & \textbf{\#Eval Views} & \textbf{PQ} \\
\midrule
3 (supervised)   & 3 & 71.37 \\
5 (3 supervised) & 3 & 71.69 \\
5 (3 supervised) & 5 & \textbf{71.85} \\
\bottomrule
\end{tabular}
\end{table}

\begin{table}[t]
\centering
\caption{\textbf{Novel-view generalization.} PQ ($\uparrow$) on MM-OR when Camera~4 is withheld during training and evaluated only at test time. Bold indicates the best result under the withheld-camera setting; the final row is a reference model trained with all cameras.}
\label{tab:missing_view}

\setlength{\tabcolsep}{6pt}
\renewcommand{\arraystretch}{1.15}
\small
\begin{tabular}{l c c c c}
\toprule
\textbf{Method} & \textbf{Cam 1} & \textbf{Cam 4 (Novel)} & \textbf{Cam 5} & \textbf{Overall} \\
\midrule
DVIS++ (w/o Cam 4)        & 54.81 & 32.43 & 53.29 & 48.53 \\
PanORama (w/o Cam 4)      & 70.65 & \textbf{50.37} & 66.97 & 62.26 \\
\midrule
PanORama (as reference)      & 72.69 & 73.03 & 69.81 & 71.85 \\
\bottomrule
\end{tabular}
\end{table}

\section{Conclusion}
In this work, we present \textbf{PanORama}, a calibration-free multiview panoptic segmentation approach for the operating room that is multiview-consistent \emph{by design}. PanORama performs cross-view feature interaction inside the backbone in a single end-to-end forward pass, allowing complementary evidence across cameras to directly inform each view’s dense prediction and yielding a globally coherent segmentation-based scene representation under sparse, partially overlapping views and strong occlusions. Our results demonstrate that PanORama achieves state-of-the-art panoptic segmentation performance on MM-OR and 4D-OR. Beyond these results, PanORama can effectively generalize to novel camera views, enabling flexible multiview configurations. By making view-consistent dense semantics feasible from sparse multiview RGB inputs, we believe PanORama paves the way for practical and scalable spatial intelligence in surgical data science.



%
%
%
%

\newpage

\bibliographystyle{splncs04}
\bibliography{content/refs}

\end{document}